\documentclass[conference]{IEEEtran}
\usepackage{cite}
\usepackage{amsmath,amssymb,amsfonts}
\usepackage{algorithmic}
\usepackage{graphicx}
\usepackage{textcomp}
\usepackage{xcolor}
\def\BibTeX{{\rm B\kern-.05em{\sc i\kern-.025em b}\kern-.08em
    T\kern-.1667em\lower.7ex\hbox{E}\kern-.125emX}}
\usepackage{subfig}
\usepackage[colorlinks=true,allcolors=blue]{hyperref}
\usepackage{multirow}
\usepackage{siunitx}
\usepackage{microtype}
\begin{document}

\title{Instantaneous Stereo Depth Estimation of Real-World Stimuli with a Neuromorphic Stereo-Vision Setup}

\author{Nicoletta Risi \qquad Enrico Calabrese \qquad Giacomo Indiveri\\
\textit{Institute of Neuroinformatics, University of Zurich and ETH Zurich, Switzerland}}

\maketitle

\begin{abstract}
The stereo-matching problem, i.e., matching corresponding features in two different views to reconstruct depth, is efficiently solved in biology. Yet, it remains the computational bottleneck for classical machine vision approaches.
By exploiting the properties of event cameras, recently proposed Spiking Neural Network (SNN) architectures for stereo vision have the potential of simplifying the stereo-matching problem. 
Several solutions that combine event cameras with spike-based neuromorphic processors already exist. However, they are either simulated on digital hardware or tested on simplified stimuli.
In this work, we use the Dynamic Vision Sensor 3D Human Pose Dataset (DHP19) to validate a brain-inspired event-based stereo-matching architecture implemented on a mixed-signal neuromorphic processor with real-world data.
Our experiments show that this SNN architecture, composed of coincidence detectors and disparity sensitive neurons, is able to provide a coarse estimate of the input disparity instantaneously, thereby detecting the presence of a stimulus moving in depth in real-time.
\end{abstract}
%%%%%%%%%%%%%%%%%%%%%%%%%%%%%%%%%%%%%%%%%%%%%%%%%%%%%%%%%%%%%%%%%%%%%%%%%%%%%
\begin{IEEEkeywords}
event-based, 3D dataset, mixed-signal hardware, analog circuits, spiking neural networks, disparity.
\end{IEEEkeywords}
%%%%%%%%%%%%%%%%%%%%%%%%%%%%%%%%%%%%%%%%%%%%%%%%%%%%%%%%%%%%%%%%%%%%%%%%%%%%%
\section{Introduction}
Depth estimation is a crucial feature in many applications, including object manipulation, surveillance, autonomous driving, and navigation. Among the various techniques explored so far, stereo vision allows retrieving 3D information by matching corresponding features in two different 2D views, i.e., by solving the \emph{stereo-matching} problem. While efficiently solved in biological systems, classical machine vision approaches require significant computational resources: Indeed, by sampling all pixels at regular time intervals, frame-based cameras suffer from data redundancy and temporal information loss. By contrast, biologically inspired neuromorphic event cameras, such as the Dynamic Vision Sensor (DVS)~\cite{lichtsteiner_128_2008_new}, transmit asynchronous streams of events generated by individual pixels in response to perceived brightness changes~\cite{gallego2019event, Posch_etal10, Berner_etal13}. 
Leveraging this sparse yet continuous encoding of visual stimuli allows to deeply simplify the stereo-matching problem. Indeed, a novel class of event-based algorithms for stereo vision, also referred to as \emph{instantaneous stereo}, extracts depth information by exploiting the inter-ocular spatio-temporal correlation of spike trains from event cameras~\cite{gallego2019event}. Moreover, since spike-based processing provides a natural interface to event-based sensing, spike-based neuromorphic hardware sets out a promising computational substrate for asynchronous, low-latency, and low-power depth estimation~\cite{steffen2019neuromorphic}. Following the pioneering work of Misha Mahowald~\cite{Mahowald94}, several Spiking Neural Networks (SNNs) that reconstruct 3D information on a per-event basis have been recently deployed on
fully digital, as well as mixed-signals neuromorphic architectures: Spinnaker~\cite{Furber_etal14, dikov2017spiking}, True North~\cite{Sawada_etal16, andreopoulos2018low}, ROLLS~\cite{Qiao_etal15,Osswald_etal17}, and DYNAP~\cite{Moradi_etal18, risi2020spike}.
Therefore, this scenario offers the remarkable opportunity to compare the same spike-based computational principles across different hardware substrates. Both~\cite{dikov2017spiking} and~\cite{andreopoulos2018low} simulate the cooperative stereo network on digital hardware. By contrast, \cite{Osswald_etal17} and~\cite{risi2020spike} use mixed-signal analog/digital neuromorphic circuits that directly emulate the dynamics of the neural computing primitives used in biology to perform stereo vision. While this approach can potentially lead to more energy-efficient and compact solutions, it suffers from noisy computation and it has been tested so far only with simplified stimuli.

Inspired by the sparse, asynchronous, and analog nature of biological computation, in this work, we approach the problem of stereo-matching with a mixed-signal neuromorphic multichip setup using a non-synthetic complex dataset.
Despite the lack of standard benchmarks for this problem domain, two datasets for event-based stereo have recently been proposed:
The Multi Vehicle Stereo Event Camera (MVSEC) Dataset~\cite{zhu2018multivehicle}, consisting of indoor and outdoor sequences recorded in a variety of illuminations and speeds, and the DVS stereo dataset~\cite{andreopoulos2018low}, with two real-world sets of sequences (a fast rotating fan and a rotating toy butterfly). 
Both datasets yield dense and high-resolution disparity maps, which make them particularly suitable for large-scale networks. While full-scale digital neuromorphic architectures of stereo vision are already available, mixed-signal neuromorphic systems are still limited to small-scale prototypes. 
Despite their small-scale (limited to a few thousand neurons per chip), preliminary estimates on the effectiveness of analog computation for event-based stereo with real-world stimuli can still be drawn from event-based datasets that yield sparse and large changes in depth. By providing DVS input data combined with precise, yet sparse, 3D ground-truth information, the DVS 3D Human Pose Dataset (DHP19)~\cite{dhp19} offers suitable samples for small-scale neuromorphic architectures of coarse stereo vision. Thus, in this work, we use the DHP19 dataset to assess the robustness of the event-based approach for neuromorphic, on-chip depth estimation recently presented in~\cite{risi2020spike}.
%------------------------------------------------------------------------
\begin{figure}[!t]
  \centering
 \subfloat[3D, mov2]{\includegraphics[trim=0 0.35in 0 0.8in,clip, width=.3\linewidth, keepaspectratio]{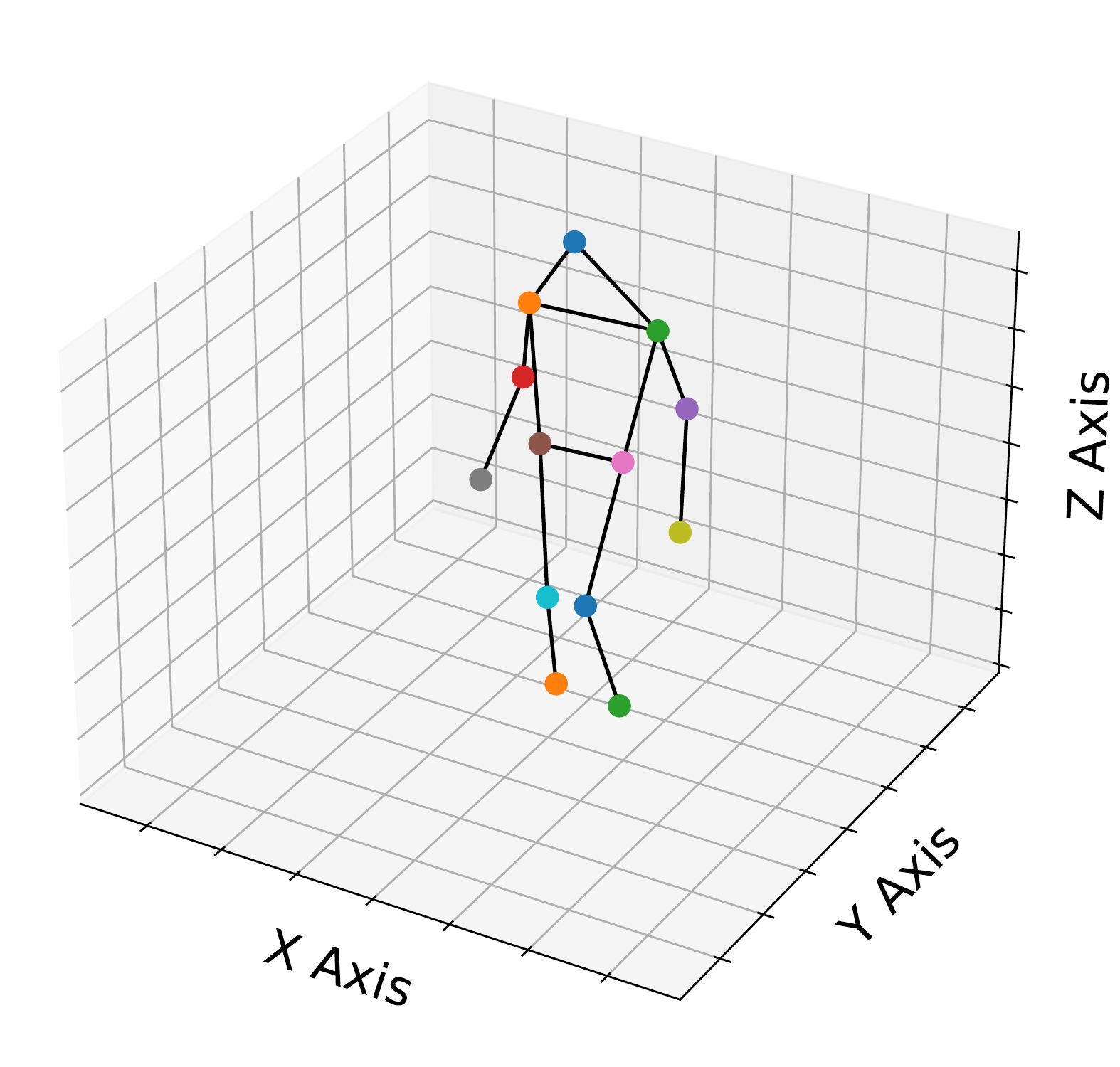}}\hspace{5pt}
 \subfloat[Left camera, mov2]{\includegraphics[width=.3\linewidth, keepaspectratio]{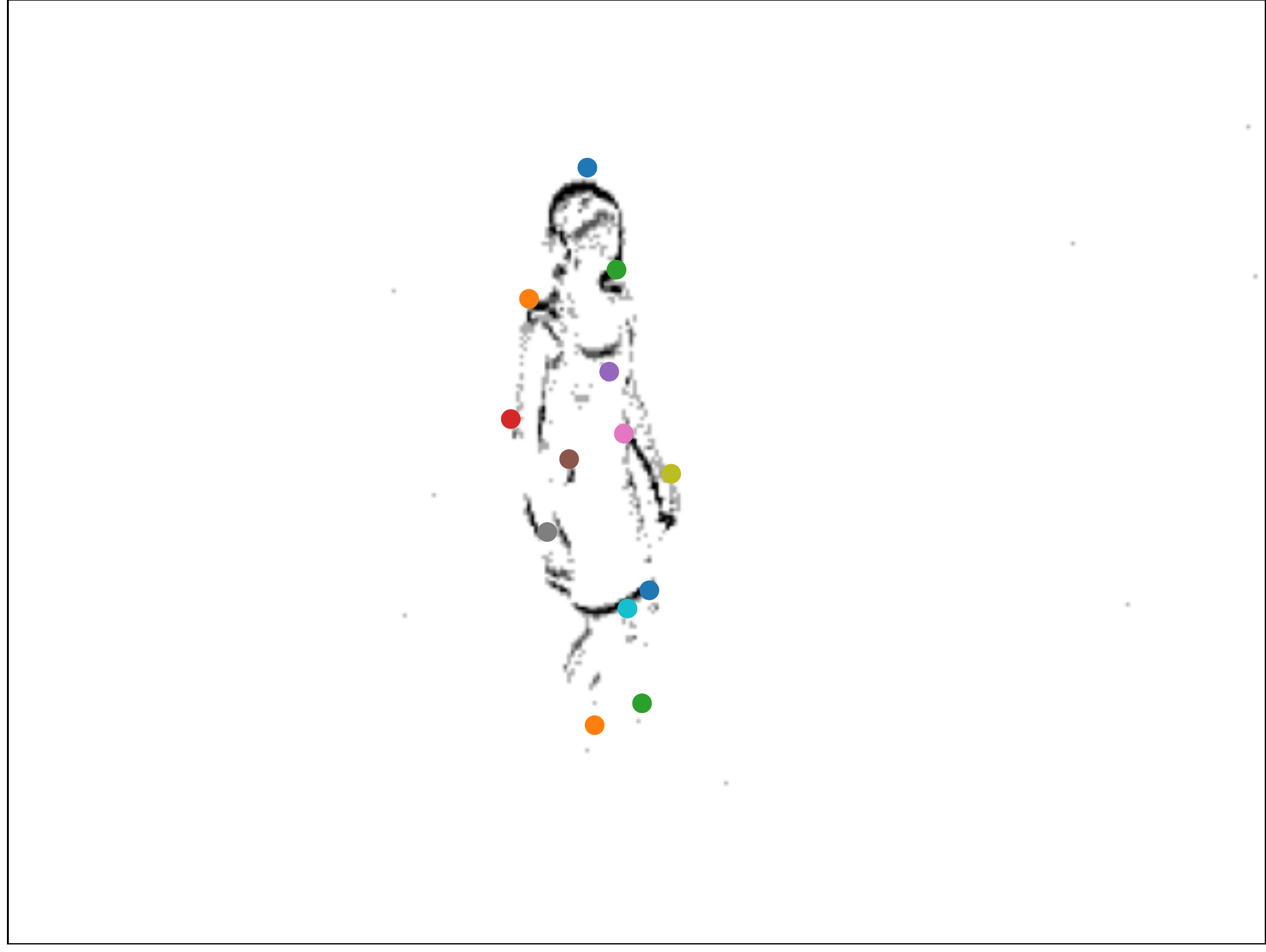}} \hspace{5pt}
 \subfloat[Right camera, mov2]{\includegraphics[width=.3\linewidth, keepaspectratio]{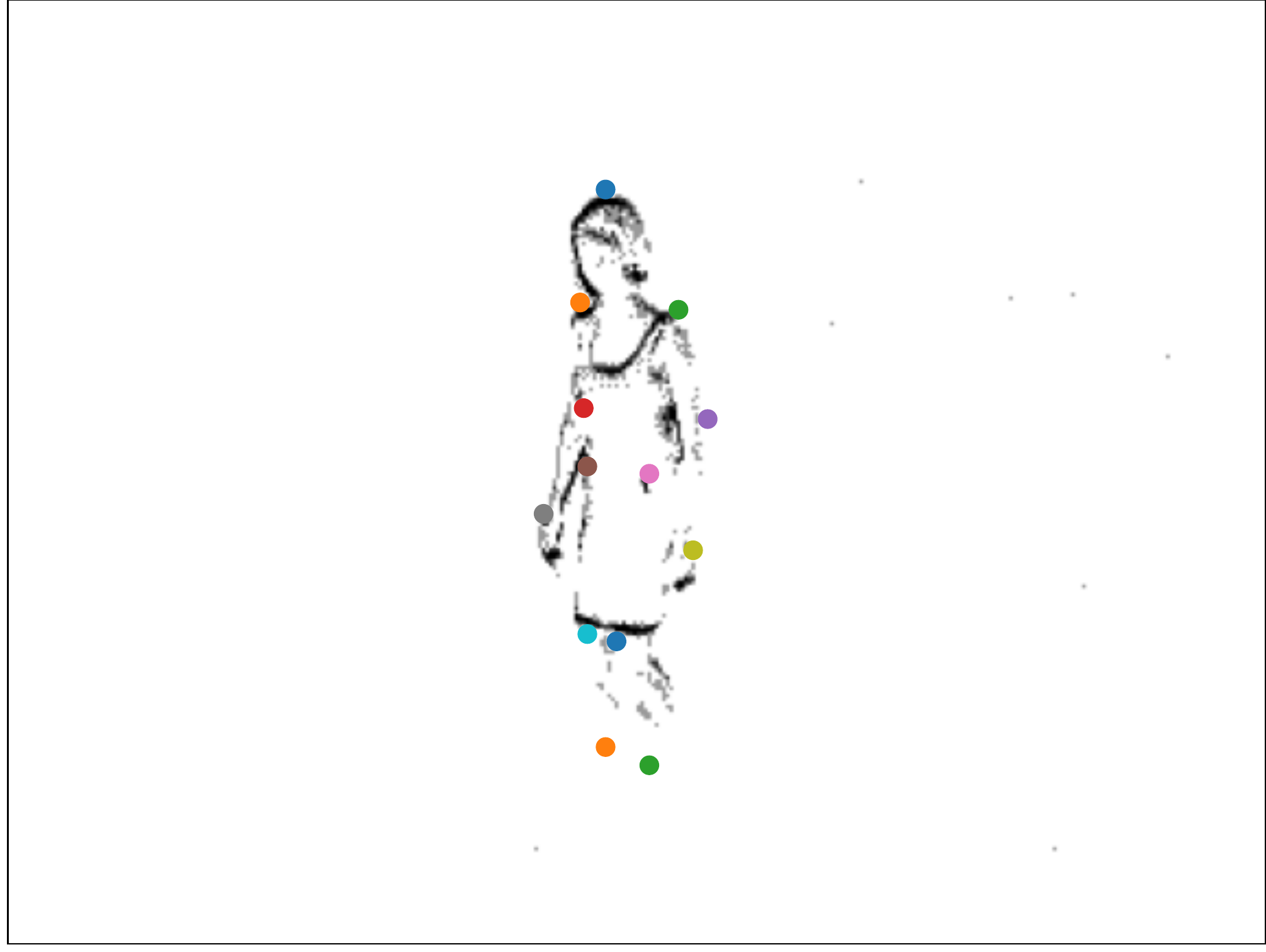}} \vspace{-5pt}
 \subfloat[3D, mov3]{\includegraphics[trim=0 0.35in 0 0.8in,clip, width=.3\linewidth, keepaspectratio]{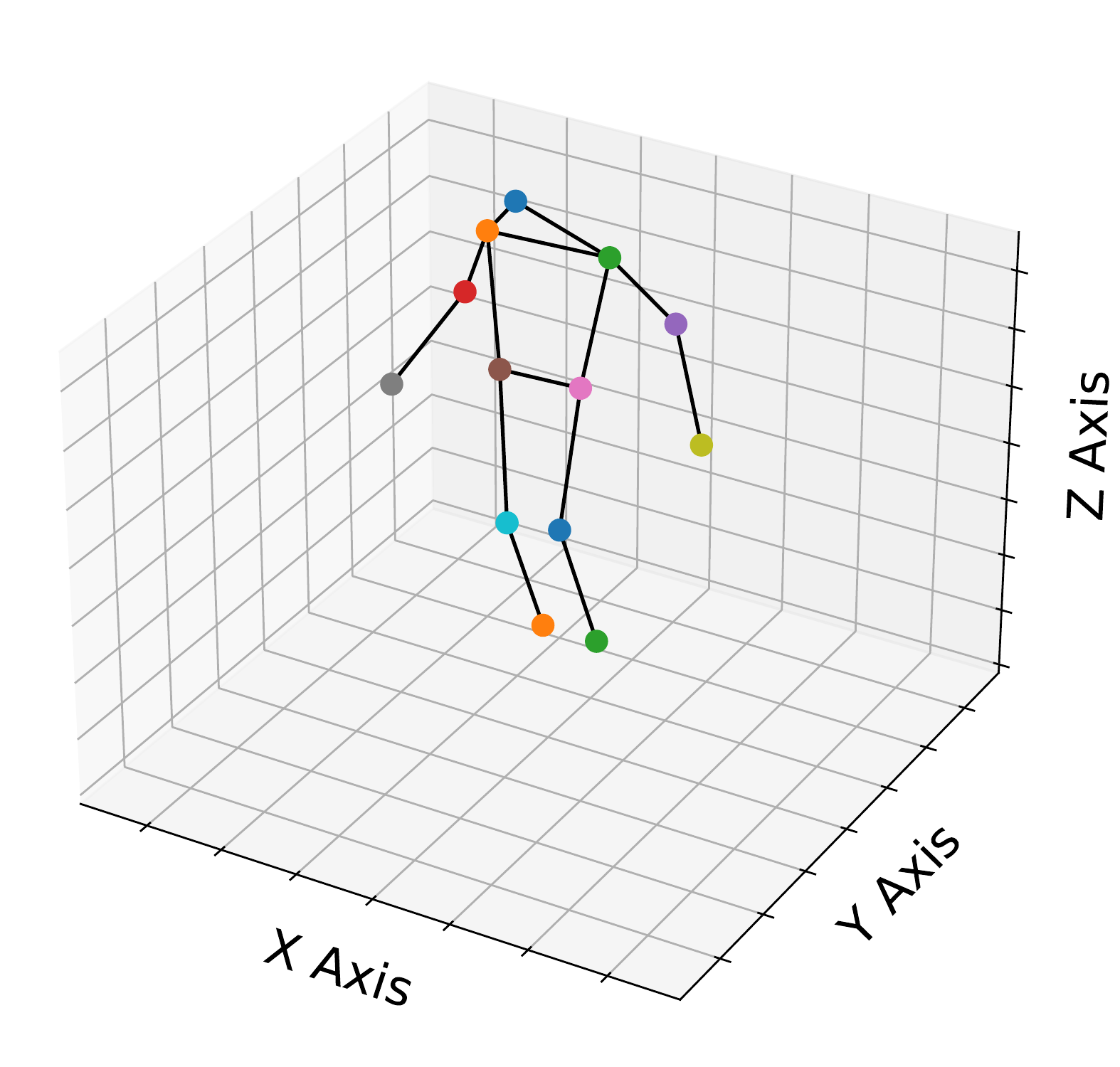}}\hspace{5pt}
 \subfloat[Left camera, mov3]{\includegraphics[width=.3\linewidth, keepaspectratio]{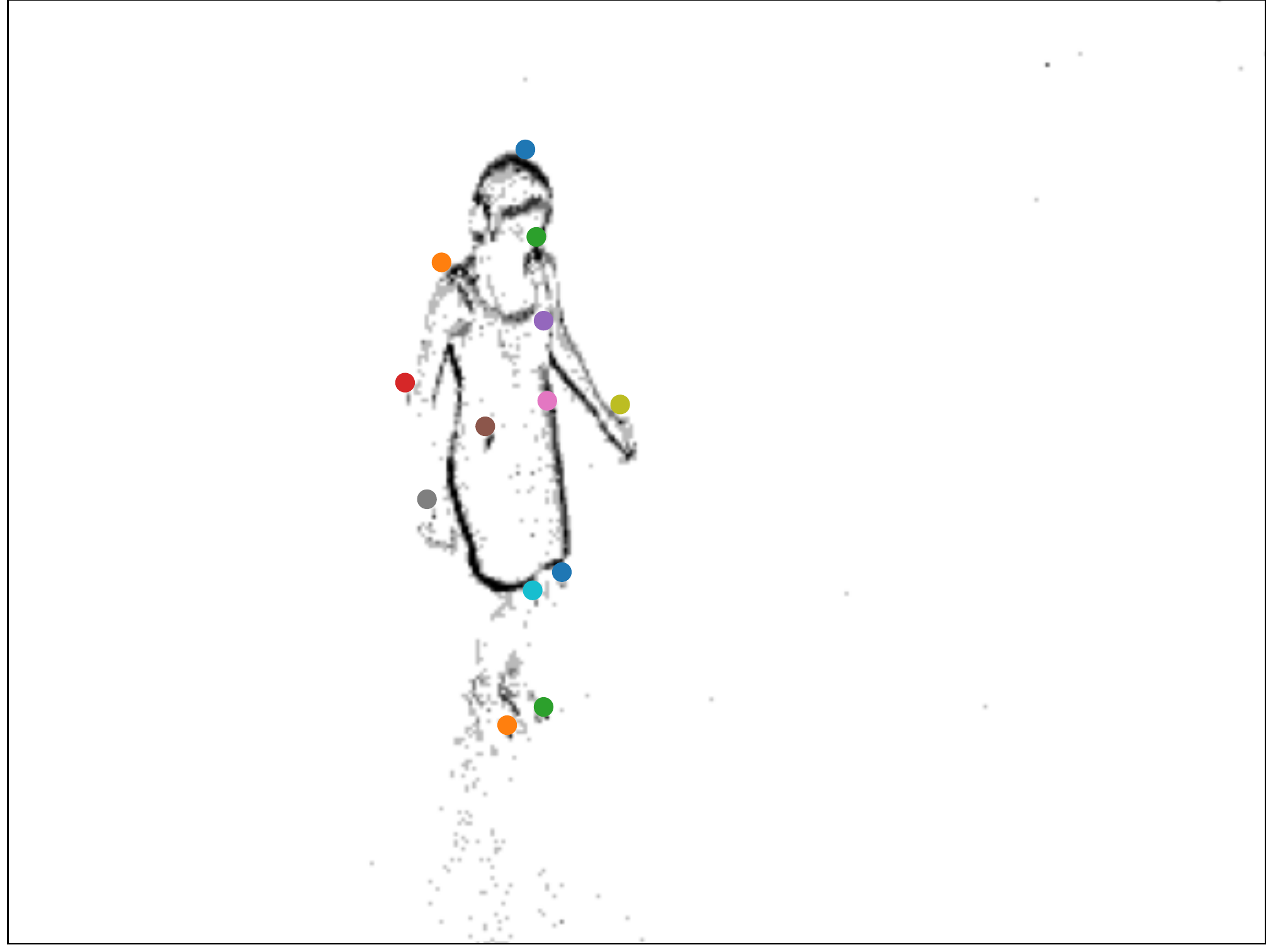}} \hspace{5pt}
 \subfloat[Right camera, mov3]{\includegraphics[width=.3\linewidth, keepaspectratio]{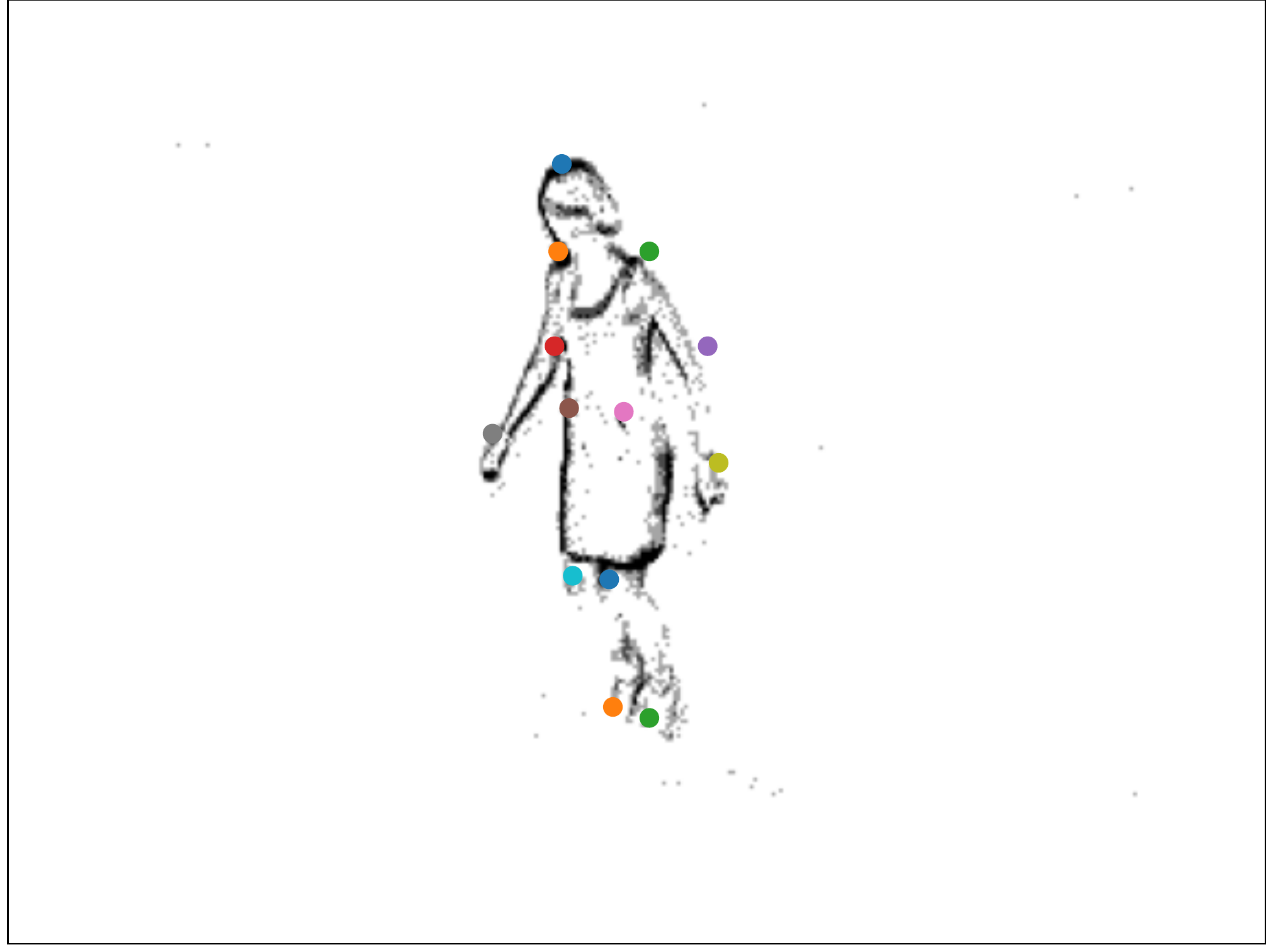}}
\caption{DHP19 Dataset samples. 3D label and 2D accumulated events with projected label for subject 1, session 2, movement 2 (first row) and movement 3 (second row).}
\label{fig:1_dataset}
\end{figure}
%------------------------------------------------------------------------

%%%%%%%%%%%%%%%%%%%%%%%%%%%%%%%%%%%%%%%%%%%%%%%%%%%%%%%%%%%%%%%%%%%%%%%%%%%%%
\section{Methodology}
\label{sec:methodology}
In this section, we introduce the event cameras, the dataset used, and the data preprocessing.
Then, we describe the SNN architecture and the neuromorphic hardware implementation.
%%%%%%%%%%%%%%%%%%%%%%%%%%%%%%%%%%%%%%%%%%%%%%%%%%%%%%%%%%%%%%%%%%%%%%%%%%%%%
\subsection{Event Cameras}
Neuromorphic event-based vision sensors are a novel class of vision sensors. Inspired by the biophysics of the retinal ganglion cells, they provide a sparse and asynchronous output of brightness-change events. The dynamic vision sensors used in this study are two DAVIS cameras such as~\cite{brandli_240x180_2014}, but with a higher resolution of $\mathrm{346\times260}$ pixels.

%%%%%%%%%%%%%%%%%%%%%%%%%%%%%%%%%%%%%%%%%%%%%%%%%%%%%%%%%%%%%%%%%%%%%%%%%%%%%
\subsection{Dataset and Data Preprocessing}
DHP19 is a dataset of human poses collected using 4 synchronized DAVIS cameras.
It is composed of recordings of 17 subjects, each performing 33 movements, and includes the 3D position of 13 joints captured using the Vicon motion capture system~\cite{ViconWebsite}.
To best assess the performance of the SNN, we selected the camera pair with the largest field of view overlap, i.e., cameras 2 and 3.
We used data from subject 1, session 2, movements 2 (\emph{single jump up-down}) and 3 (\emph{single jump forwards}), which have different depth changes as seen from the two cameras. 
Specifically, movement 2 is characterized by a small depth change, while movement 3 has a larger depth change.
Figure~\ref{fig:1_dataset} shows data from the DHP19 subset used in our experiments. 
The advantage of using the DHP19 dataset is that it combines sparse recordings from event cameras with high spatial resolution 3D information, providing the ground-truth 3D position of the markers without further processing required by machine vision algorithms or parameter tuning. 

\begin{figure}[!t]
    \centering
    \includegraphics[trim=0 0.05in 0.1in 0.1in,clip,width=0.95\columnwidth]{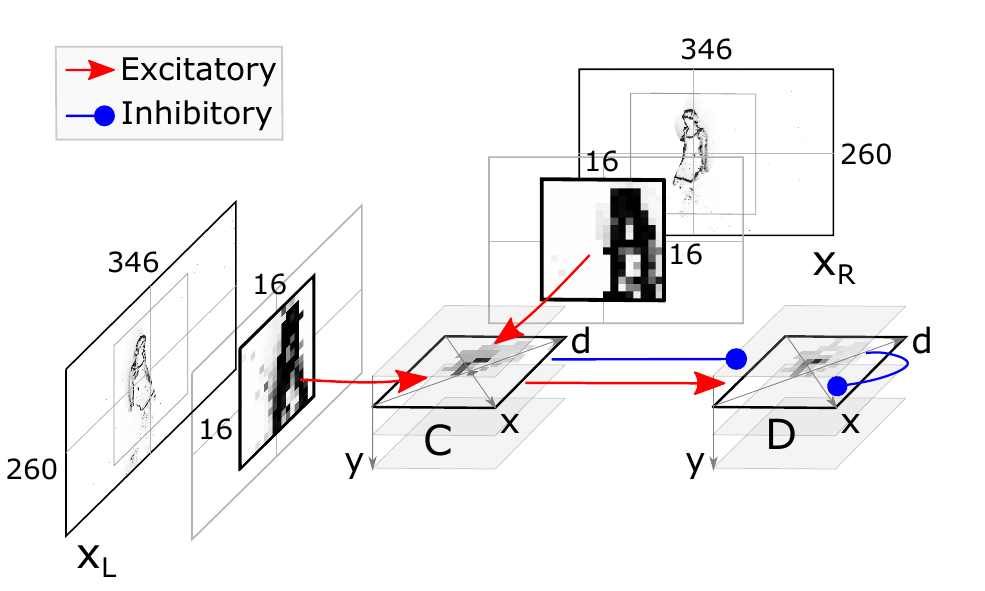}
    \caption{SNN architecture scheme. Events from the two DAVIS cameras are downscaled and sent to the SNN, composed of a coincidence population (C) and a disparity population (D). See~\cite{risi2020spike} for a comprehensive description of the architecture.}
    \label{fig:arch}
\end{figure}

The raw event streams are preprocessed to filter out noise events and to reduce the camera output resolution, in order to fit the constraints imposed by the neuromorphic processor.
The noise filtering on the events is done following the schedule proposed in~\cite{dhp19}: Background noise and hot pixels are removed, and events due to Vicon cameras infrared light are masked.
The output resolution of each DAVIS camera is reduced by first applying a uniform downscaling, where regions of $\mathrm{6\times6}$ non-overlapping pixels are mapped to a single pixel,
then a crop of $\mathrm{16\times16}$ pixels is extracted as the input for the neuromorphic hardware SNN (\figurename~\ref{fig:arch}).
%------------------------------------------------------------------------
%------------------------------------------------------------------------

%%%%%%%%%%%%%%%%%%%%%%%%%%%%%%%%%%%%%%%%%%%%%%%%%%%%%%%%%%%%%%%%%%%%%%%%%%%%%
\subsection{The Spike-Based Neuromorphic Architecture}
The spike-based neuromorphic architecture used to extract disparity information
is based on the hardwired topology proposed in~\cite{risi2020spike}. It consists of a SNN of event-based stereo vision emulated on three multicore analog/digital Dynamic Neuromorphic Asynchronous Processors (DYNAP)~\cite{Moradi_etal18} integrated in a 4-chip board. Input visual streams from the DHP19 dataset are sent to the neuromorphic processor via a dedicated Field Programmable
Gate Array (FPGA) device (Xilinx Kintex-7 FPGA on the OpalKelly XEM7360), which supports SuperSeed USB3.0 data transfer. In the next sections, we introduce the SNN model and the neuromorphic processor. 
%%%%%%%%%%%%%%%%%%%%%%%%%%%%%%%%%%%%%%%%%%%%%%%%%%%%%%%%%%%%%%%%%%%%%%%%%%%%%
\subsubsection{The Spiking Neural Network}\label{subsec:SNN}
The SNN architecture of event-based cooperative stereo vision, shown in \figurename~\ref{fig:arch}, is adapted from the structure presented in~\cite{Osswald_etal17, risi2020spike}.
It consists of three neuronal populations: The \emph{retina} cells, emulated by two downscaled $\mathrm{16\times16}$ pixels of the DAVIS cameras, and two 3D arrays of Leaky Integrate and Fire (LIF) silicon neurons, with $\mathrm{N=2\times1024}$ \emph{coincidence} (C) neurons (grouped into two sub-populations of excitatory and inhibitory ones) and $\mathrm{N=1024}$ \emph{disparity} (D) neurons.
Each coincidence and disparity neuron is assigned a triplet of coordinates, which determine the neuron representation of a location in 3D space: A horizontal cyclopean position $x_n = x_{R} + x_{L}$, a vertical cyclopean position $y_n = y_R = y_L$, and a disparity value $d_n = x_{R} - x_{L}$.
Each neuron in the retina cells targets, via excitatory connections, neurons in the coincidence population that are tuned to its same spatial location ($x_{R}$ or $x_{L}$). Coincidence neurons are tuned to respond to temporally synchronized interocular events only, thereby implementing coincidence detection. However, as temporal information is crucial but not enough to effectively solve the correspondence problem, the spiking activity within population C encodes all potential binocular stereo matches. This ambiguity is solved in the disparity population by means of inhibitory and excitatory connections from the coincidence neurons: Each disparity neuron receives feed-forward inhibitory inputs from all coincidence neurons tuned to the same cyclopean position and excitatory inputs from all coincidence neurons tuned to the same disparity. Recurrent inhibition across disparity neurons tuned to the same line of sight (i.e., $x=x_{L}$ or $x=x_{R}$) enforces competition across potential binocular matches. As shown in~\cite{Osswald_etal17}, this connectivity scheme effectively implements the matching constraints of cooperative stereo algorithms (uniqueness and continuity), with disparity neurons approximating the local covariance of the binocular inputs.

%%%%%%%%%%%%%%%%%%%%%%%%%%%%%%%%%%%%%%%%%%%%%%%%%%%%%%%%%%%%%%%%%%%%%%%%%%%%%
\subsubsection{DYNAP Neuromorphic Processor}
The SNN model of event-based cooperative stereo vision is emulated on three four-core asynchronous mixed-signal neuromorphic processors, the DYNAP~\cite{Moradi_etal18}, fabricated using standard \SI{0.18}{\micro\metre} 1P6M CMOS technology.
Each core comprises 256 Adaptive Exponential Integrate-and-Fire (AEI\&F) silicon neurons that emulate the biophysics of their biological counterparts, and four different dedicated analog circuits that mimic fast and slow excitatory/inhibitory synapse types~\cite{Brette_Gerstner05}. Each neuron has a Content Addressable Memory (CAM) block, containing 64 programmable entries allowing to customize the on-chip connectivity. A fully asynchronous inter-core and inter-chip routing architecture allows flexible connectivity with microsecond precision under heavy system loads. Digital peripheral asynchronous input/output logic circuits are used to receive and transmit spikes via an Address Event Representation (AER) communication protocol~\cite{Deiss_etal98}.

%%%%%%%%%%%%%%%%%%%%%%%%%%%%%%%%%%%%%%%%%%%%%%%%%%%%%%%%%%%%%%%%%%%%%%%%%%%%%
\subsection{Neuromorphic Architecture Performance}\label{subsec:data_analysis}
The 3D information from the Vicon motion capture system was used as ground truth. First, the 3D positions of the 13 joints were projected to the 2D camera planes. Then, the projected coordinates were mapped to each downscaled camera view according to the scaling factor applied to the input events. The resulting marker locations in the two camera views were used to obtain a spatially coarse and uniformly sampled ground-truth disparity trajectory across time $d_V$. By contrast, the stimulus disparity encoded by the SNN was defined as the firing-rate weighted average of the encoded disparity $d_n$ for each neuron in C and D, or population Center of Mass (CoM)~\cite{dayan2001theoretical}:
\begin{equation}
    CoM[t_i] = \frac{\sum_{n}^{N} r_n[t_i] d_n}{\sum_{n}^{N}r_n[t_i]}, 
    \label{eq:CoM}
\end{equation}
 with $r_n$ being the neuron $n$ instantaneous firing rate, sampled at discrete time steps $t_i$.
 
To quantify the architecture performances, two metrics were used to compare the SNN output with the Vicon ground truth:\\
$\bullet$ Root Mean Square Error (RMSE) between the SNN $CoM$ and the Vicon disparity $d_V$.\\
$\bullet$ Percentage of Correct Disparities (PCD), defined as:
    \begin{equation}
        PCD = \sum_{i} \frac{TD[t_i]}{FD[t_i]+TD[t_i]}
    \end{equation}
with $TD[t_i]$ and $FD[t_i]$ being True and False Disparity events. In each time window $t_i$, spikes were labelled as TD if generated by neurons encoding for: $$d_n \in \left[ \min(d_V[t_i]) - \epsilon_d , \max(d_V[t_i]) + \epsilon_d \right] $$
with $\epsilon_d=1$.

Finally, for each input sample, we estimated the power consumption of the mixed-signal neuromorphic implementation as described in~\cite{risi2020spike}.

\section{Experimental Results}\label{sec:results}
%------------------------------------------------------------------------
\begin{table}[t]
    \centering
    \caption{Architecture Performance - DHP19 samples.}
    \label{table:results_}
    \begin{tabular}{c|l|l|c|c|c}
    \hline
    \multirow{3}{*}{Subject} & \multirow{3}{*}{Session} & \multirow{3}{*}{Movement} & \multicolumn{2}{c|}{Metric}  & Est. Power \\ \cline{4-5} 
                             &                          &                           & PCD          & \multirow{2}{*}{RMSE}          & Consumption  \\ 
                             &                          &                           & $(\epsilon_d = 1)$          &           &  [uW] \\ 
    \hline
    \hline
    1                        & \multicolumn{1}{c|}{1}   & \multicolumn{1}{c|}{2}    &  0.98        & 0.70          & 18.9  \\ 
    \hline
    1                        & \multicolumn{1}{c|}{1}   & \multicolumn{1}{c|}{3}    &  0.99        & 2.01          & 25.7 \\ 
    \hline
\end{tabular}
\end{table}

%-----------------------------------------------------------------------
\begin{figure*}[!t]
\centering
\fboxsep=1pt
\subfloat[]{\fbox{\begin{minipage}{0.48\textwidth}\centering\includegraphics[width=.95\textwidth,trim=200 0 205 0, clip]{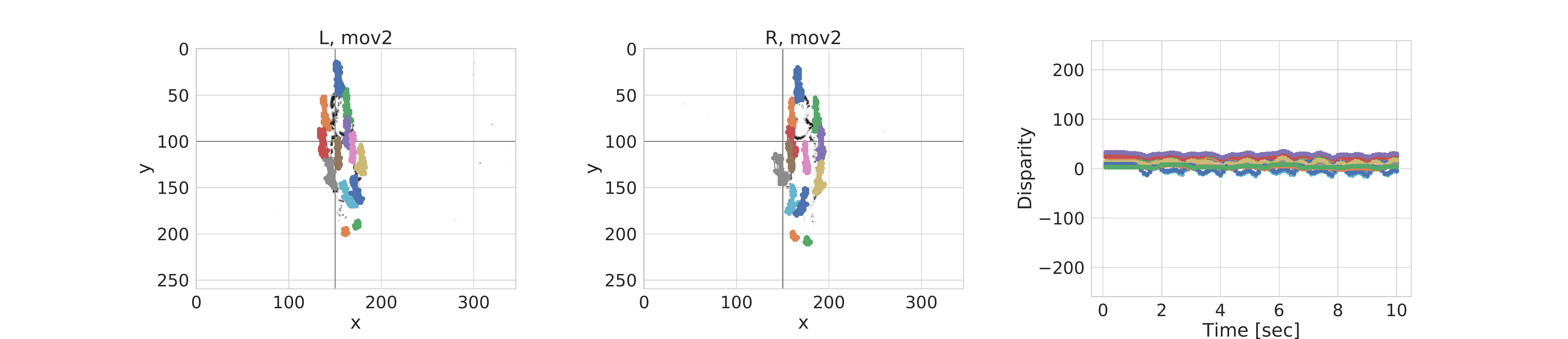}
\includegraphics[width=1\textwidth,trim=200 0 205 20 ,clip]{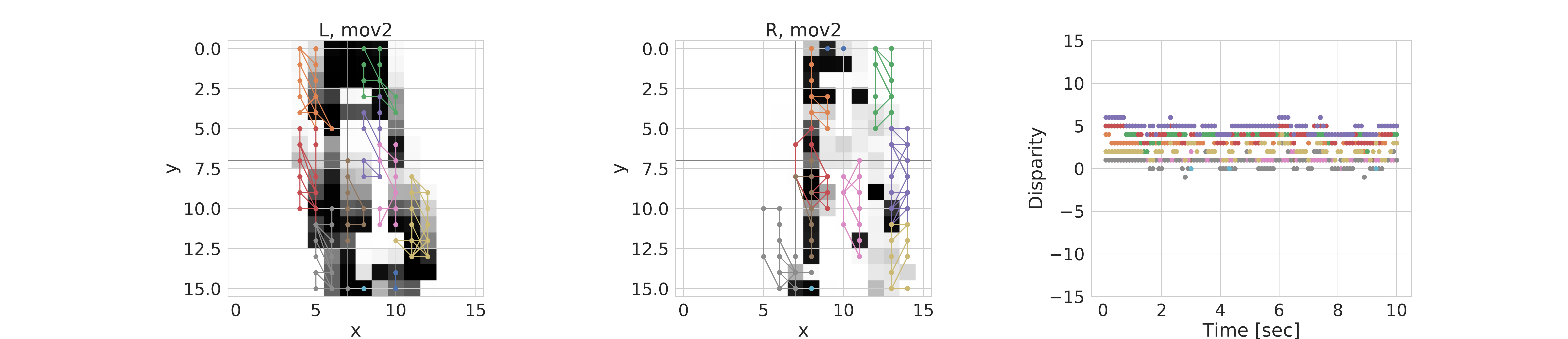}
\vspace{1pt}
\end{minipage}}}\hfill
\subfloat[]{\fbox{\begin{minipage}{0.48\textwidth}\centering\includegraphics[width=.95\textwidth,trim=200 0 205 0, clip]{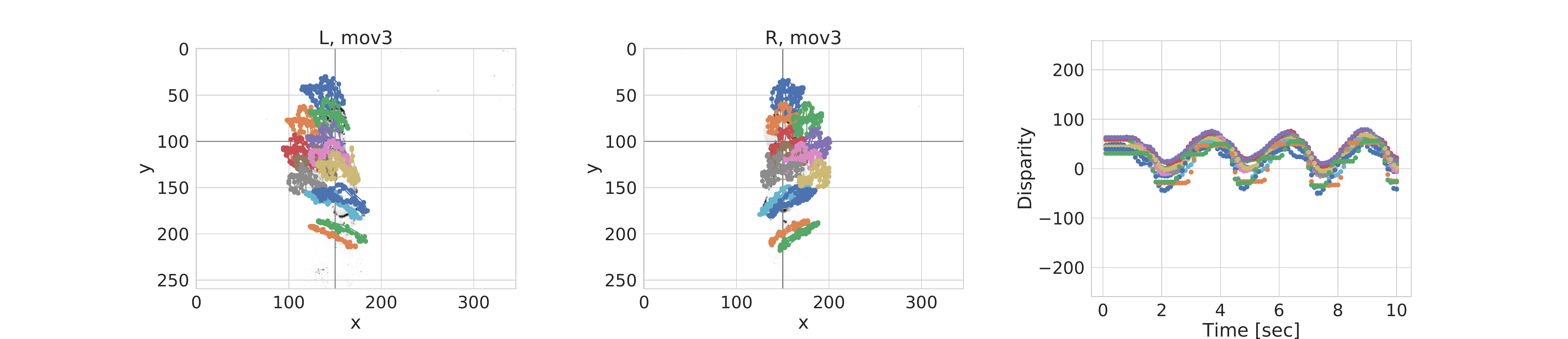}
\includegraphics[width=1\textwidth,trim=200 0 205 20, clip]{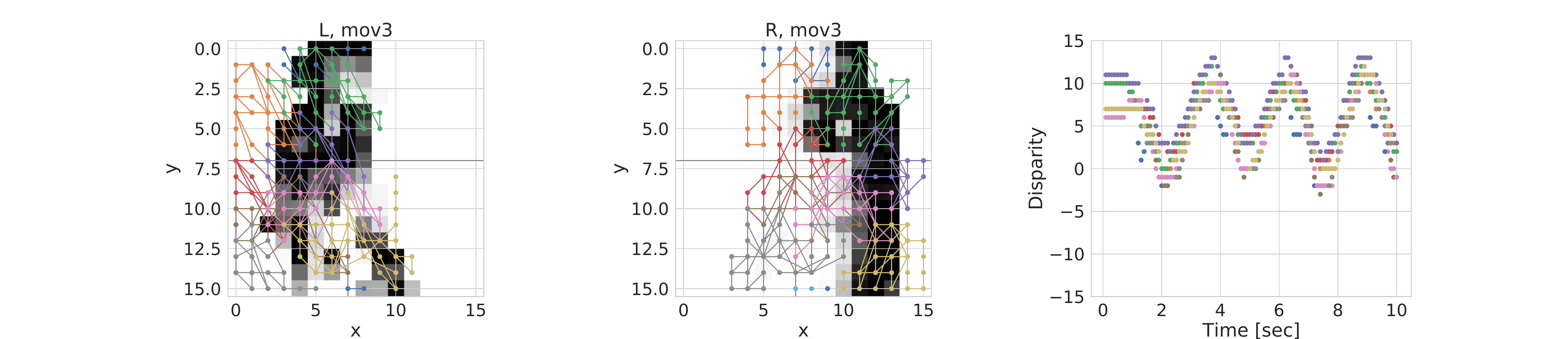}
\vspace{1pt}
\end{minipage}}}
\vspace{3pt}
\includegraphics[width=.7\textwidth,trim=5 5 5 0,clip]{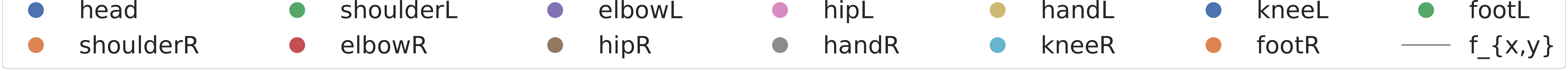}
\caption{Full resolution (top) and reduced resolution (bottom) SNN input and markers disparity for movement 2 (a) and 3 (b).}
\label{fig:3_input_and_disparity}
\end{figure*}
%------------------------------------------------------------------------
\begin{figure*}[t]
  \centering
\subfloat[]{\fbox{\includegraphics[width=0.48\textwidth,trim=80 0 150 0, clip]{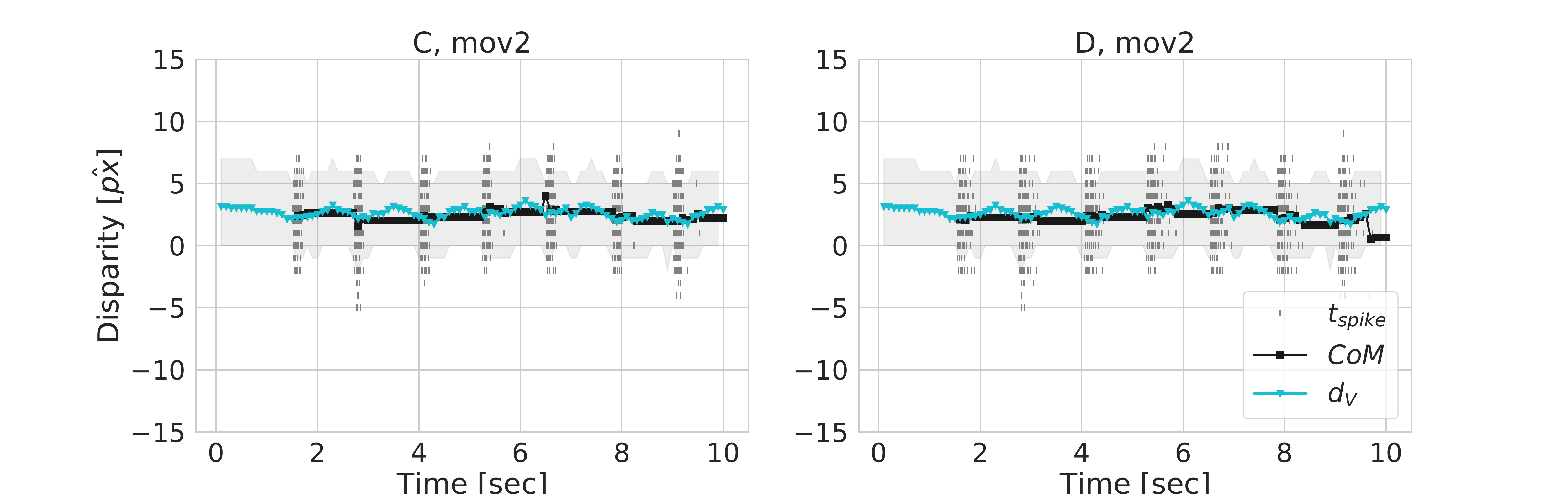}}}\hfill
\subfloat[]{\fbox{\includegraphics[width=0.48\textwidth,trim=80 0 150 0, clip]{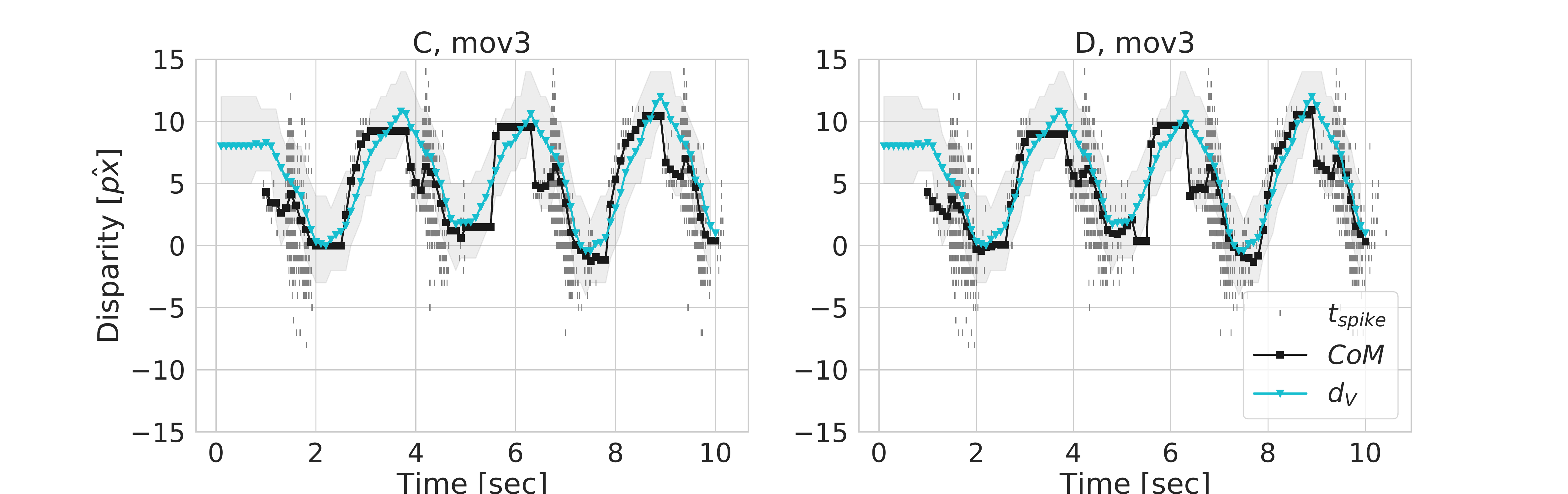}}}
\caption{Raster plot of coincidence and disparity neurons, population $CoM$ and average markers disparity $d_V$ for movement 2~(a) and 3~(b). Neuron \emph{id}s are sorted with respect to their associated disparity value, expressed in downscaled pixels ($\hat{px}$). The shaded area represents the range of TD spikes ($\epsilon_d=1$).}
\label{fig:4_rasterplot_time}
\end{figure*}
\begin{figure*}[t]
\centering
\subfloat[]{\fbox{\includegraphics[width=.48\textwidth, keepaspectratio]{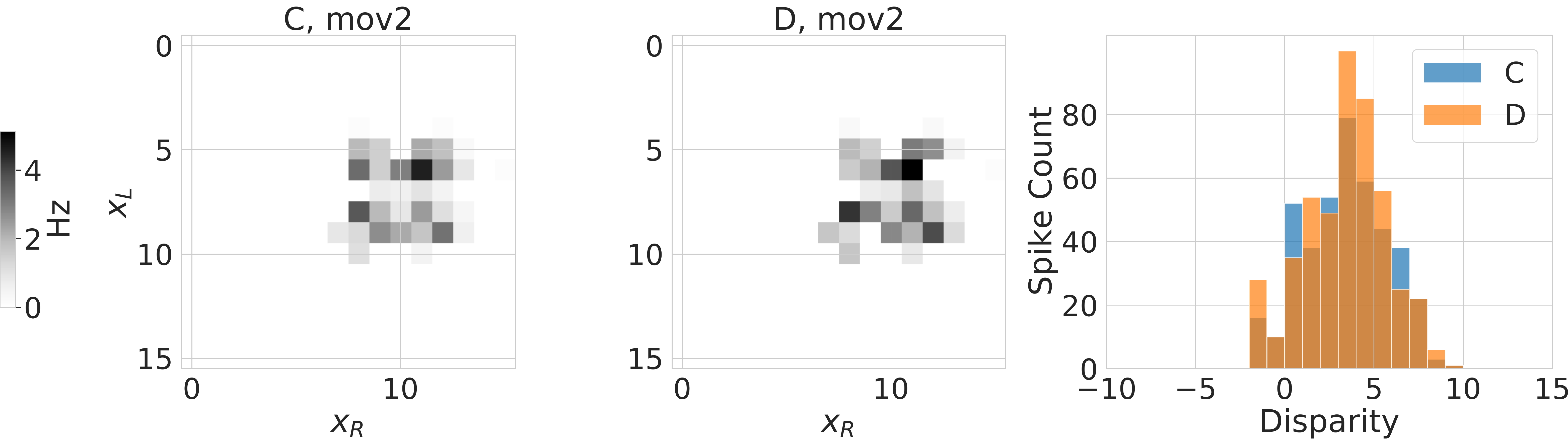}}}\hfill
\subfloat[]{\fbox{\includegraphics[width=.48\textwidth,
keepaspectratio]{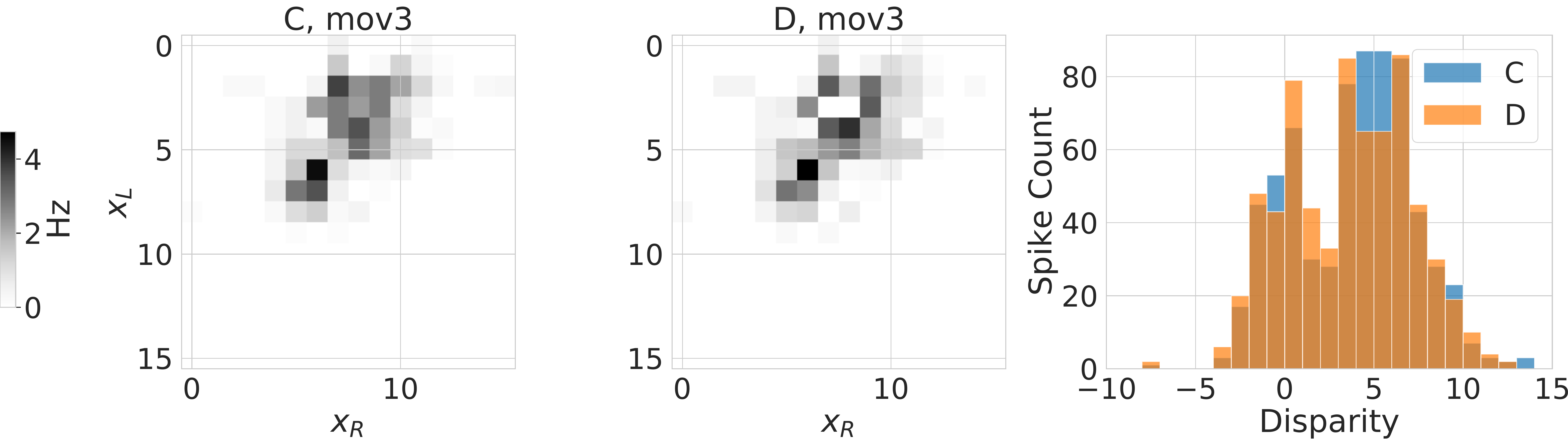}}}
\caption{Mean firing rate of neurons tuned to the same cyclopean position $y_n$ from coincidence (left) and disparity (center) populations, and histogram of encoded disparity values (right) for movement 2~(a) and 3~(b).}
\label{fig:4_mfr}
\end{figure*}

%------------------------------------------------------------------------
Figure~\ref{fig:3_input_and_disparity} shows the events from movement 2 (\figurename~\ref{fig:3_input_and_disparity}a) and 3 (\figurename~\ref{fig:3_input_and_disparity}b) of the DHP19 dataset. For each column, the top row shows the events of both cameras, depicted as time surfaces~\cite{lagorce2016hots} with rectified polarities, together with the projected marker locations, and their corresponding disparity over time. The same representation is used to depict the reduced resolution input data (bottom row), fed to the SNN. 
The marker disparities are significantly different across the two movements, for both full and reduced resolution data, reflecting the changes in stimulus depth across time.
Figure~\ref{fig:4_rasterplot_time} shows the spiking activity in time of both C and D populations, with neuron \emph{id}s sorted with respect to their associated disparity values. When compared to the ground truth $d_V$, the population $CoM$ shows that the firing rate of the silicon neurons can effectively provide a real-time coarse estimation of the input disparity. 
Figure~\ref{fig:4_mfr} shows the mean firing rate measured in a subset of C and D neurons tuned to the same cyclopean position $y_n$. As opposed to movement 2 (\figurename~\ref{fig:4_mfr}a), movement 3 (\figurename~\ref{fig:4_mfr}b) elicits neural activity spread along the main diagonal of the 2D arrays of neurons, which comprises units tuned to different disparity values, and therefore signals the presence of a stimulus moving in depth. This is also reflected by the histogram of encoded disparities, with movement 3 eliciting activity in a wider disparity range. Table~\ref{table:results_} shows the obtained values of RMSE and PCD, and the estimated power consumption in both samples.

%------------------------------------------------------------------------
%------------------------------------------------------------------------

%%%%%%%%%%%%%%%%%%%%%%%%%%%%%%%%%%%%%%%%%%%%%%%%%%%%%%%%%%%%%%%%%%%%%%%%%%%%%
\section{Discussion and Conclusion}
In this work, we demonstrate the feasibility of coarse, low-power depth estimation of real-world stimuli using event-based, mixed-signal neuromorphic hardware. 
Given their massively parallel, asynchronous, real-time computing features, and their explicit representation of time and space~\cite{Indiveri_Sandamirskaya19}, as opposed to their fully-digital time-multiplexing counterpart, these systems have the potential to achieve higher energy-efficient computation in a reliable way, despite the inherent noise and variability in their individual neural CMOS circuits.
The major contribution of this work is the validation of a neural architecture for coarse, real-time, stereo vision with events from real-world stimuli. Unlike datasets such as~\cite{andreopoulos2018low} and~\cite{zhu2018multivehicle}, which can be used to compute the ground truth on a per-event basis, the Vicon marker-based motion capture system provides ground-truth depth information directly with sparse data linked to point labels attached to specific body parts. This approach is therefore suited for validating current small-scale analog implementations of event-based stereo vision and provides a compelling benchmark for cross-platform comparisons. While additional analysis of more samples from the DHP19 dataset can be useful for a full characterization of our event-based stereo-vision setup, this work sets the stage for using the proposed approach to validate novel low-power, coarse depth estimation systems that could be deployed in applications ranging from robotics to surveillance.
%------------------------------------------------------------------------
\section*{Acknowledgments}
This work was supported by the ERC Grant NeuroAgents (Grant No. 724295).

\bibliographystyle{ieeetr}
\bibliography{biblio/biblio}
%------------------------------------------------------------------------
\end{document}